# Global Wheat Challenge 2020: Analysis of the competition design and winning models

Authors: Etienne David, Franklin Ogidi, Wei Guo, Frederic Baret, Ian Stavness

## 1. Abstract

Data competitions have become a popular approach to crowdsource new data analysis methods for general and specialized data science problems. In plant phenotyping, data competitions have a rich history, and new outdoor field datasets have potential for new data competitions. We developed the Global Wheat Challenge as a generalization competition to see if solutions for wheat head detection from field images would work in different regions around the world. In this paper, we analyze the winning challenge solutions in terms of their robustness and the relative importance of model and data augmentation design decisions. We found that the design of the competition influence the selection of winning solutions and provide recommendations for future competitions in an attempt to garner more robust winning solutions.

## 2. Introduction

Crowdsourcing is an increasingly popular approach for scientists to make advances in their field by collecting diverse raw or labelled data (Gao, Barbier and Goolsby, 2011; Prill *et al.*, 2011; Wiggins and Crowston, 2011; Giuffrida *et al.*, 2018), solving problems that are difficult for algorithms but easy for humans, such as protein folding (Koepnick *et al.*, 2019), or accessing large-scale distributed computing power (E. J. Korpela *et al.*, 2011). Crowdsourcing of data analysis has increased rapidly in recent years due to the popularity of Big Data challenges (Yang *et al.*, 2018) on web platforms such as Kaggle or Codalab. In particular, problems that are amenable to machine learning approaches, such as computer vision problems, have been promoted and popularized through data competitions such as ImageNet or COCO (Deng *et al.*, 2009; Lin *et al.*, 2014).

Data competitions have also expanded to specific application areas, such as image-based plant phenotyping, where deep learning methods have been employed for plant disease classification (Albetis et al., 2017; Fuentes et al., 2017; Toda and Okura, 2019), plant and organ detection and counting (Madec et al., 2019; David et al., 2020; Ayalew, Ubbens and Stavness, no date), vegetation segmentation (Mortensen et al., 2016). In this context, data competitions help to connect domain experts, e.g. plant scientists, with computer and data scientists. Such collaboration with data scientists outside the traditional scope of plant phenotyping is essential to solve fundamental data science problems within the domain.

Early computer vision competitions for plant phenotyping problems (Scharr *et al.*, 2016; Giselsson *et al.*, 2017) such as leaf counting ((Minervini *et al.*, 2016)) and leaf segmentation ((Scharr *et al.*, 2016)) were highly successful. Plant phenotyping competitions have typically focused on indoor, controlled plant images although a few recent competition have included outdoor field data such as CVPR Agriculture-Vision competition (Chiu *et al.*, 2020) or the GrassClover competition(Skovsen *et al.*, 2019) . Expanding data competition agricultural applications are crucial for tackling new challenges faced by global food production due to climate change pressures. The availability of agricultural sensors, vectors, and data is rapidly expanding, but by effective data processing algorithms remains a limitation. Therefore, crowdsourcing new approaches could help to address data analysis challenges in plant phenotyping and agriculture.

Despite the use of deep learning in recent plant phenotyping studies, the robustness and generalizability of these methods, particular for small plant datasets with limited sample version, remains an open question. The issue of robustness in trained plant phenotyping models has been difficult to study due to a lack of real-world datasets (Geirhos *et al.*, 2020), including in plant phenotyping (David *et al.*, 2021). A large and diverse dataset is required to study the robustness problem. Most of the Deep Learning algorithms require a large set of labelled images to be trained on, such as ImageNet (Deng *et al.*, 2009) or MS COCO (Lin *et al.*, 2014). The process of labelling such images can be long and tedious, limiting the accessibility to various datasets. The presence of diversity in the datasets is also important to study the robustness of solutions by having realistic bias. Efforts to collect and clean datasets to make them accessible are still new (Koh *et al.*, 2021). Large datasets in plant phenotyping exist for in-door conditions (Minervini *et al.*, 2016; Tsaftaris and Scharr, 2019) but few exist with more than one location and for outdoor conditions.

Localizing wheat head in field conditions is an interesting problem to study as a model: the potential leverage in terms of accuracy of yield estimation resulting to improved wheat research, and the lessons learned could be replicated to other use cases such as other plants or organ localization and/or counting, leaf area estimation, root segmentation etc. Several papers proposed successful implementations (Madec *et al.*, 2019; Xiong *et al.*, 2019) for wheat head detection but only on a limited range of variability. It is unclear if their results expand to new datasets. Variation of sensors, illumination conditions, development stages and genotype could impact the performance as described in (David *et al.*, 2021). The algorithms themselves are prone to error from small variations (Rosenfeld, Zemel and Tsotsos, 2018).

We introduced the Global Wheat Head Dataset (GWHD) (David *et al.*, 2020) in order to study the robustness problem in plant phenotyping and as a challenge to crowdsource new solutions. Localizing wheat heads in field image is an appropriate model problem (Madec *et al.*, 2019; Xiong *et al.*, 2019), the lessons learned from which could be replicated for other use cases such as other plants or organ localization and/or counting, leaf area estimation, root segmentation etc. The GWHD is a labelled dataset for wheat head localization containing 4700 images and 185k annotated wheat heads from 9 institutions across 7 countries. We design the problem by creating a bias between the training and the test sets. The bias is based on geography, as a proxy for a genotype bias. Sub datasets from Europe and North America were put in the training dataset and sub datasets from Asia and Oceania (Australia) were put in the test set.

Based on the GHWD, we hosted a data competition on Kaggle called the Global Wheat Challenge 2020 (GWC 2020) to attract a large cohort of ML practitioners to solve the wheat head localization problem with a general method. The competition took place from 4th May to 4th August and attracted up to 2245 competitors. During a Kaggle competition, models from different competitors are evaluated on a test set with a unique metric. At the time of the competition, the metric was the accuracy at different IoU levels. The 3 best models on this metric are meant to be made open source. Models proposed by competitors are often a mix of models with hyperparameter engineering and are the result of the combination of lots of ideas. During the competition, several original data augmentation strategies were proposed and were supposed to be the key to the winning solutions according to their authors.

In this paper, we provide a summary of the competition, including the approached employed by the top three winning solutions. We provide a detailed analysis of these solutions to analyze their robustness and evaluate their performance on a new test set of unseen wheat images. We also provide an ablation study of the most popular architecture used (EfficientDet) and the most popular data augmentation strategies in order elucidate the mechanisms that contributed to model performance and robustness on the test set. Finally, we discuss lessons learned from hosting the competition and provide recommendations for future competitions.

# 3. Material and Methods
## 3.1. Datasets

| Kaggle Name | GWHD Name | Owner | Ccountry | Location | Date | Labeled images |
|---|---|---|---|---|---|---|
| Ethz_1 | ethz_1 | ETHZ | Switzerland | Usask | 06/06/2018 | **747** |
| Rres_1 | rres_1 | Rothamsted | UK | Rothamsted | 13/07/2015 | **432** |
| Arvalis_1 | arvalis_11 | Arvalis | France | GLB | date_1 | **66** |
|  | arvalis_12 | Arvalis | France | GLB | date_2 | **401** |
|  | arvalis_13 | Arvalis | France | GLB | date_3 | **588** |
| Arvalis_2 | arvalis_2 | Arvalis | France | GLB | 27/05/2019 | **204** |
| Arvalis_3 | arvalis_31 | Arvalis | France | VLB | 06/06/2019 | **448** |
|  | arvalis_32 | Arvalis | France | VSC | 26/06/2019 | **160** |
| Inrae_1 | inrae_1 | INRAe | France | Toulouse | 28/05/2019 | **176** |
| Usask_1 | usask_1 | USaskatchewan | Canada | Saskatchewan | 06/06/2018 | **200** |
| utokyo_1 | utokyo_11 | UTokyo | Japan | Loc 1 | 22/05/2018 | **538** |
|  | utokyo_12 | UTokyo | Japan | Loc 1 | 22/05/2018 | **456** |
| utokyo_2 | utokyo_3 | UTokyo | Japan | Loc 2 | Multiple date | **120** |
| Ukyoto_1 | ukyoto_1 | UKyoto | Japan | Kyoto | 30/04/2020 | **60** |
| NAU_1 | NAU_1 | Nanjing Agricultural University | China | Baima | 1 | **20** |
| UQ_1 | uq_11 | UQueensland | Australia | Gatton | 12/08/2015 | **22** |
| UQ_1 | uq_12 | UQueensland | Australia | Gatton | 08/09/2015 | **16** |
| UQ_1 | uq_13 | UQueensland | Australia | Gatton | 15/09/2015 | **14** |
| UQ_1 | uq_14 | UQueensland | Australia | Gatton | 01/10/2015 | **30** |
| UQ_1 | uq_15 | UQueensland | Australia | Gatton | 09/10/2015 | **30** |
| UQ_1 | uq_16 | UQueensland | Australia | Gatton | 14/10/2015 | **30** |

**Table 1: Characteristics of the datasets from Global Wheat Challenge 2020 and the adaptation from the Kaggle name to the revised subdatasets.**

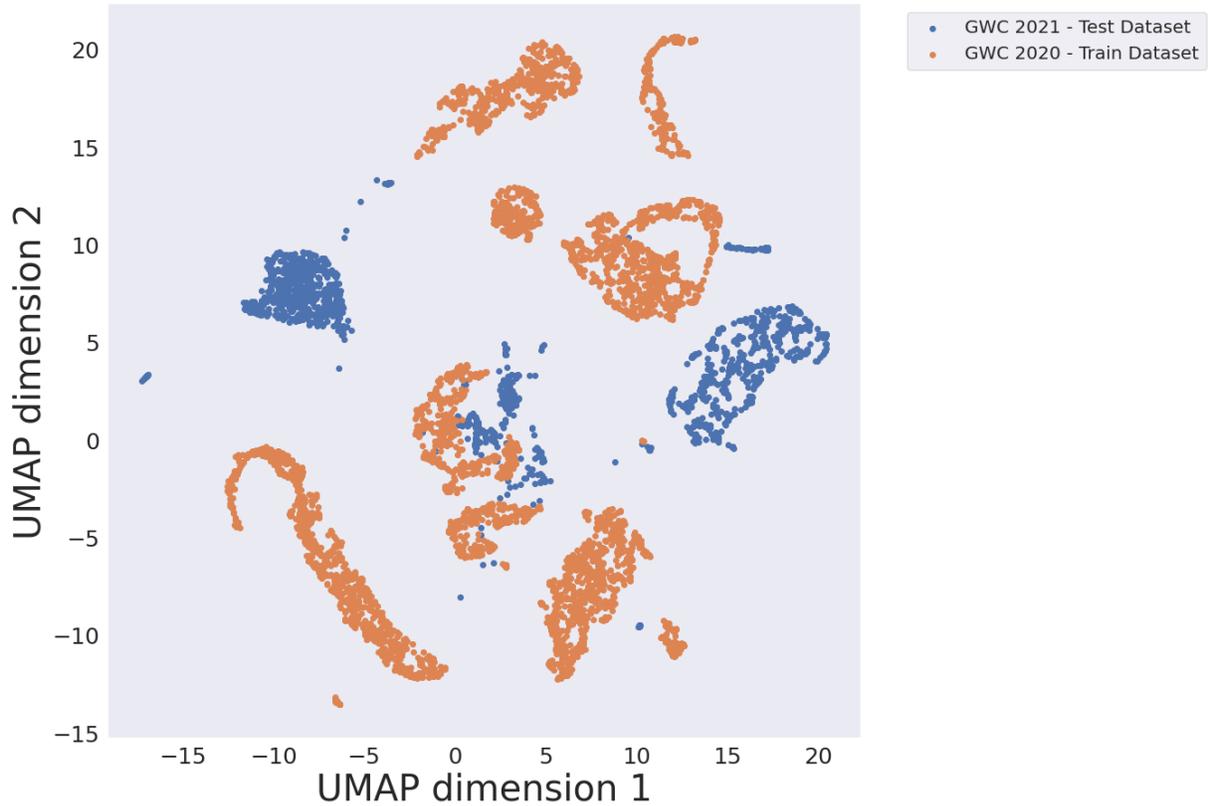

**Figure 1: Visualization of the dataset diversity with pre-trained encoding and UMAP projection**

The dataset used for the competition is the Global Wheat Head Detection dataset (GWHD) (David *et al.*, 2020). GWHD is designed to solve the challenge of wheat head localization. The main challenge is to train an algorithm that is robust to variation of sensors, illumination, and organ morphology. A total of 4700 images containing 19000 wheat heads from 9 institutions across 4 continents and 7 countries are included in GWHD. Their characteristics are described in Table 1. Our objective is to quantify how far the Kaggle competition has advanced the wheat head localization task. Competitors have worked for 3 months to optimize their performance on 4 sub datasets: utokyo_1, utokyo_2, nau_1, uq_1. Each one of these sub datasets can be considered as a new domain. A more formal definition of subdataset has been introduced in the new Global Wheat Head Dataset (David et al., paper in preparation). Some subdatasets used during the Kaggle competition are then split: for instance UQ_1 is now represented as 6 different subdatasets. If there is no specific mention, subdatasets used during the competition are used in this paper. Due to the low number of test domains, it is very likely that the whole wheat head diversity is not entirely covered. The competitors are likely to have overfit on these domains rather than solving the robustness problem. We therefore use the new datasets from the Global Wheat Head Dataset 2021 (David et al. , paper in preparation): it represents 26 new subdatasets, including new images from countries such as Belgium, Norway, USA, Mexico and Sudan.

Multiple feedbacks from competitors revealed labelling errors, therefore the dataset have been corrected after the competition. The latest version is openly available on Zenodo. The annotations used for the Kaggle challenge will be called the "Kaggle labels" and the one published on Zenodo, called the "Zenodo labels". Both sets of annotations will be used in the following study. A qualitative analysis of the introduced diversity has been conducted. A 512D feature vector is extracted for each image from all sub datasets, including the new ones, with a network (VGG 16) pre-trained on ImageNet. The UMAP (McInnes, Healy and Melville, 2020) algorithm is used for dimensionality reduction. The additional sub datasets seem to cluster apart from the sub datasets used during the GWC 2020. These visual differences introduce new bias, unseen during the GWC 2020.

### 3.2. Presentation of winning solutions

The output of the competition are three open-source models. It comes in the form of trained weights and Jupyter notebooks to reproduce the inference process. The Jupyter notebooks are exported from a "Kaggle kernel", a cloud-based notebook allowing competitors to use cloud computing resources. The competitors did not have access to the test image and had to apply pseudo-labelling blindly. We reorganized the code in different files to separate pseudo-labelling from predictions, but we did not alter the winning solutions. The three winners come from various background: Vietnam, US and Slovenia. They were invited to a one-hour virtual interview to explain their solutions and the rationale behind their process.

During the competition design, we expected competitors to try to improve domain-adaptation approaches with advanced data augmentation strategies, domain-adversarial training, and were tentative for new architectures that could solve the problem of overlapping wheat head and small wheat heads. Surprisingly, all winners used existing open-source architectures such as EfficientDet, Faster-RCNN and Yolo-v3, without any specific domain adaptation module. The use of different architectures indicates that more than one architecture can generalize to different datasets and, as such, the network architecture is not the main factor in performance. Additionally, one part of the performance comes from test-time augmentation combined the weighted boxes fusion (Solovyev, Wang and Gabruseva, 2019). Although this is an expected result, it is something often overlooked in plant phenotyping.

During post-competition interviews, winners reported that Data augmentation and pseudo labelling were key in their performance. Winners all used pseudo-labelling. Pseudo-labelling (Lee, 2013) is the practice of converting prediction over the test set to label called "pseudo-label" and fine-tuning the model with a mix of training data and pseudo-labeled data. Winners used several data augmentation techniques such as Mixup (Zhang *et al.*, 2018) and Mosaic augmentation, an augmentation technique described in YoloV4 (Bochkovskiy, Wang and Liao, 2020). Additional strategies were developed during competition. A popular strategy was to solve the "jigsaw puzzle". Given that some images were cropped from the same, larger images, competitors have recreated large images and used them to re-crop new images randomly instead to apply a regular grid as in the preprocessing stage like in (David *et al.*, 2020) . This strategy, while interesting, is out of the scope from our original objective and therefore did not had been investigated further. It is however difficult to assess the effect of the specific data augmentation strategies on robustness. We identified this specific topic as a question to investigate, with the goal of finding a better baseline strategy than Faster-RCNN (used in (David *et al.*, 2020) ) .

| Rank | Solution name | Data Augmentation | Architecture | Inference | Kaggle Score |
|---|---|---|---|---|---|
| 1 | DungNB | Mixup ; Custom mosaic augmentation | EffDet (7x) + FasterRCNN | TTA + WBF | 0.6897 |
| 2 | OverFeat | **Mixup, Cutmix** | Effdet X 2 | TTA + WBF | 0.6879 |
| 3 | Javu | Mixup | YoloV3 +DarkNet53 | TTA + WBF | 0.6839 |

**Table 3: Summary of winning solutions**

### 3.3. Data augmentation ablation study: Design and implementation details

The 20 most-voted python notebooks from the Kaggle competition were reviewed to select data augmentation strategies to analyze. About 40 unique augmentation techniques were used across these kernels. Most of these augmentation strategies are applied only to a single image, while some others require a batch of images to be mixed in interesting ways. All the multi-image augmentation techniques used across the top notebooks, including Mixup (Zhang *et al.*, 2018), Cutmix (Yun *et al.*, 2019) and mosaic, were selected for further analysis. About 15 single-image augmentation techniques were selected for further analysis by grouping the techniques and dropping ones that were too similar in function (e. g. GlassBlur vs. GaussianBlur) or ones that were deemed unnecessary for the purpose of the analysis (e.g. RandomSnow or RandomSunFlare). Only one augmentation per group of similar operations is selected randomly for each batch. The single-image data augmentation techniques were implemented using a popular image processing library known as Albumentations.

The multi-image augmentation techniques (Mixup, Cutmix) were used from Overfeat solution (Table 3, Figure 2). Mixup (Zhang *et al.*, 2018) and Cutmix (Yun *et al.*, 2019) both mix 2 images to form an augmented image. While Mixup uses a weighted sum of the pixels from the two images, Cutmix replaces a patch in one image with a randomly cut patch from another image. In this study the weights for the Mixup operation were set to a constant, 0.5, although it could also be drawn from a random distribution for each image. The Mosaic augmentation technique is an extension of Cutmix and takes a batch of 4 images, randomly scales each image and patches them together into a single image. Because of its close nature to cutmix, it has not be included in the ablation study. The annotations for each image are proportionately scaled and stacked together to form the augmented annotations. The EfficientDet-D4 (Tan, Pang and Le, 2020) object detection model, pretrained on the Microsoft Common Objects in Context (MS-COCO) dataset (Lin *et al.*, 2014) was selected as the baseline model for evaluating the effects of these data augmentation techniques on the wheat detection task using the GWHD dataset. All data augmentation strategies used in this study were applied only to the training dataset, with a probability of 0.5.

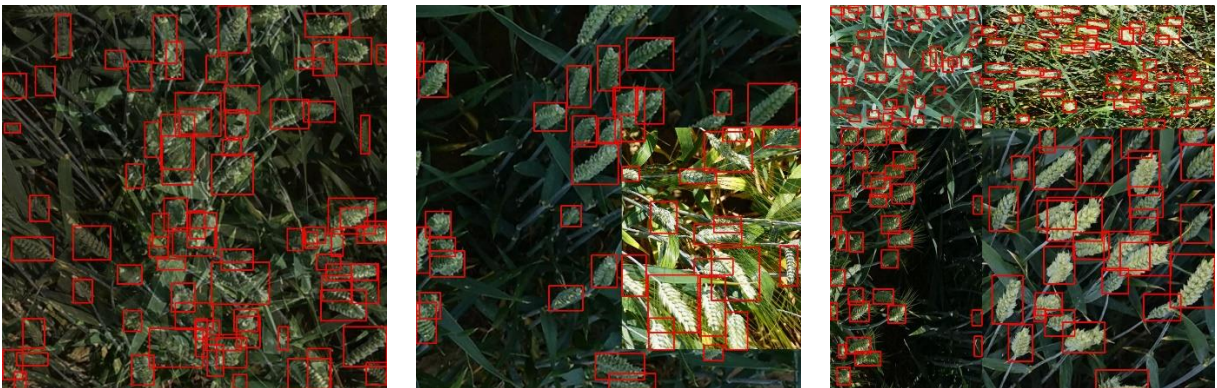

**Figure 2: Presentation of Mixup, Cutmix and Mosaic strategy**

### 3.4. Evaluation Metrics

The goal of the dataset and the competition is to find a general solution to the wheat head localization problem. The Intersection over Union (IoU) is used to define a True Positive (TP) ratio. It is the ratio of intersection between two surfaces over their union. A TP is a ground truth bounding box that matches with a predicted bounding box over an IoU threshold. A False Positive (FP) is a predicted bounding box that does not match with any ground truth bounding box. A False Negative (FN) is a ground truth bounding box that does not match with any predicted bounding box. Accuracy is the number of TP on the sum of TP, FP and FN, described in equation 1

$$Accuracy_{image} = \frac{TP}{TP + FP + FN}$$

The metric used for Kaggle competition was called "Average Precision" in the context of the competition but will be called "Accuracy" in this paper to disambiguate with the common accepted definition of Average Precision in computer vision(Everingham *et al.*, 2010). The accuracy for each image is calculated at different IoU threshold levels, from 0.5 to 0.75 with step of 0.05. The "Mean Accuracy" is the average of accuracies.

The metric used for Kaggle is computed globally, without any weighting on the number of images per dataset. There is a risk that the winning solutions are skewed towards a better performance on the larger dataset (utokyo_1) than finding generic solutions. We propose to weight the accuracy depending on the weight of each subdataset. We call this metric the "Weighted Accuracy". We additionally use the confusion matrix term to benchmark the different models.

$$Accuracy_{Kaggle} = \frac{1}{n} * \sum_{i=1}^{n} Accuracy_i$$

$$Weighted\ Accuracy = \frac{1}{D} \sum_{d=1}^{D} \frac{1}{n_d} * \sum_{i=1}^{n_d} Accuracy_{di}$$

4. Results and Discussion
    4.1. Limits of Kaggle competition design

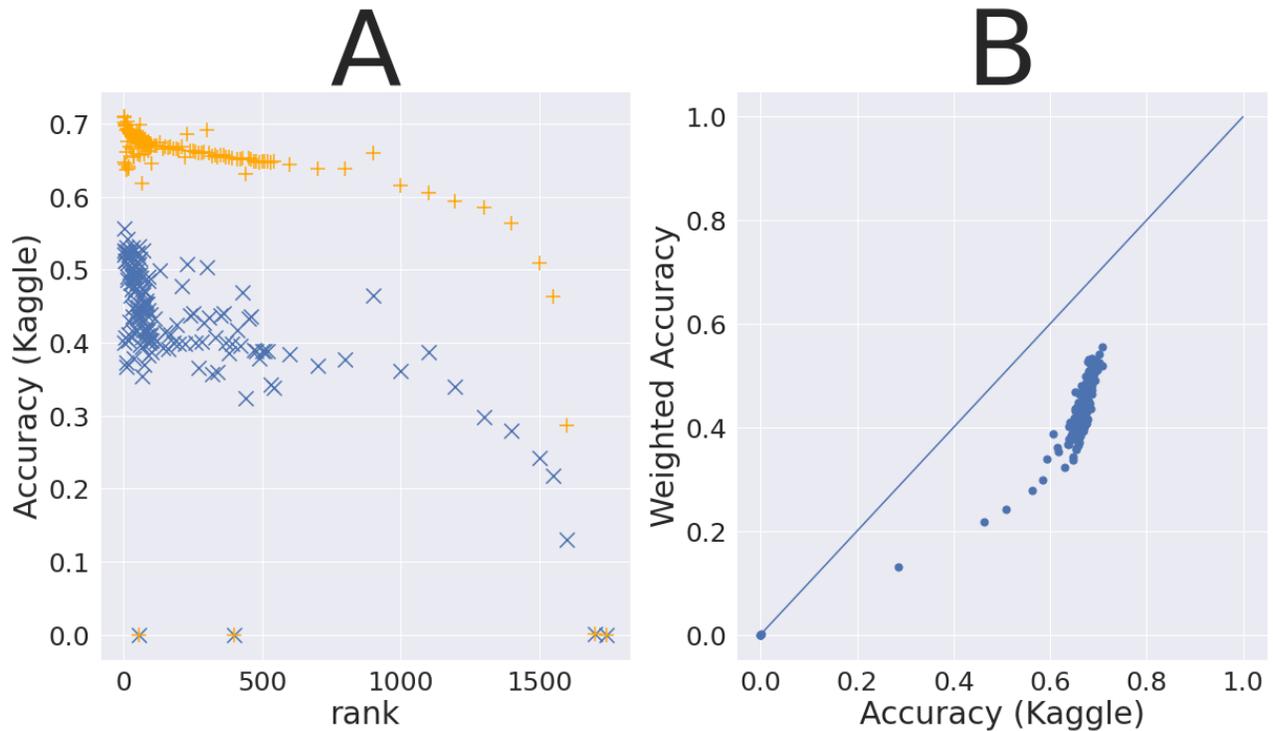

Figure 3: Presentation of the limit of the ranking used during the competition. The x-axis of Part A represents the actual ranking in the Kaggle competition while the y-axis shows the score. The scores with our re-implementation of the metric are shown in orange and the proposed simplified metric are shown in blue. Both are evaluated on the corrected private test set. Part B the score for each solution with metric used during Kaggle against the ADA. Part C presents a simulation of the new ranking based on the new score and limited to the solutions sampled for the analysis.

The design of a competition is critical to obtain solutions that satisfy its objective. The metric proposed during the Kaggle competition presents the drawback of not having an open implementation. Our open re-implementation from scratch reaches similar but unidentical scores. These small changes can lead to severe changes in ranking. The results presented in figure 3-A reflect how the corrections and the re-implementation could have drastically changed the ranking on the Kaggle annotations.

Domain generalization is a core problem that the dataset aims to solve but the metric chosen for the Kaggle challenge was too heavily influenced by the performance on the largest domain – utokyo_1. The use of weighted accuracy promotes solutions that have more balanced performance. Figure 3-B displays how the weighted accuracy is less saturated than the original metric with performance varying between 0.3 and 0.6 while the AA varies between 0.65 and 0.7.

Applying these results would have strongly influenced the final ranking. [Praxis](Overfeat), which has ranked 2nd and VinBigDataMedical (DungNB) which has ranked 1st would have ranked 1st and 3rd, respectively. Peculiarly, the solution that was ranked 9th would be 2nd, and the 3rd place solution would have dropped below a rank of 100. These results demonstrate the robustness of the solutions by Praxis and VinBigDataMedical despite the original metric. However, the former metric could have discouraged even more robust solutions from rising to the top. The solution by Praxis will be studied closely in the rest of the paper because it scored significantly higher (+0.03) than the second solution on the weighted accuracy. It will be called the "Kaggle solution" in the rest of the study. Henceforth, the score will be computed with the weighted accuracy on the whole public and private test set.

## 4.2. Analysis of best model performance improvement

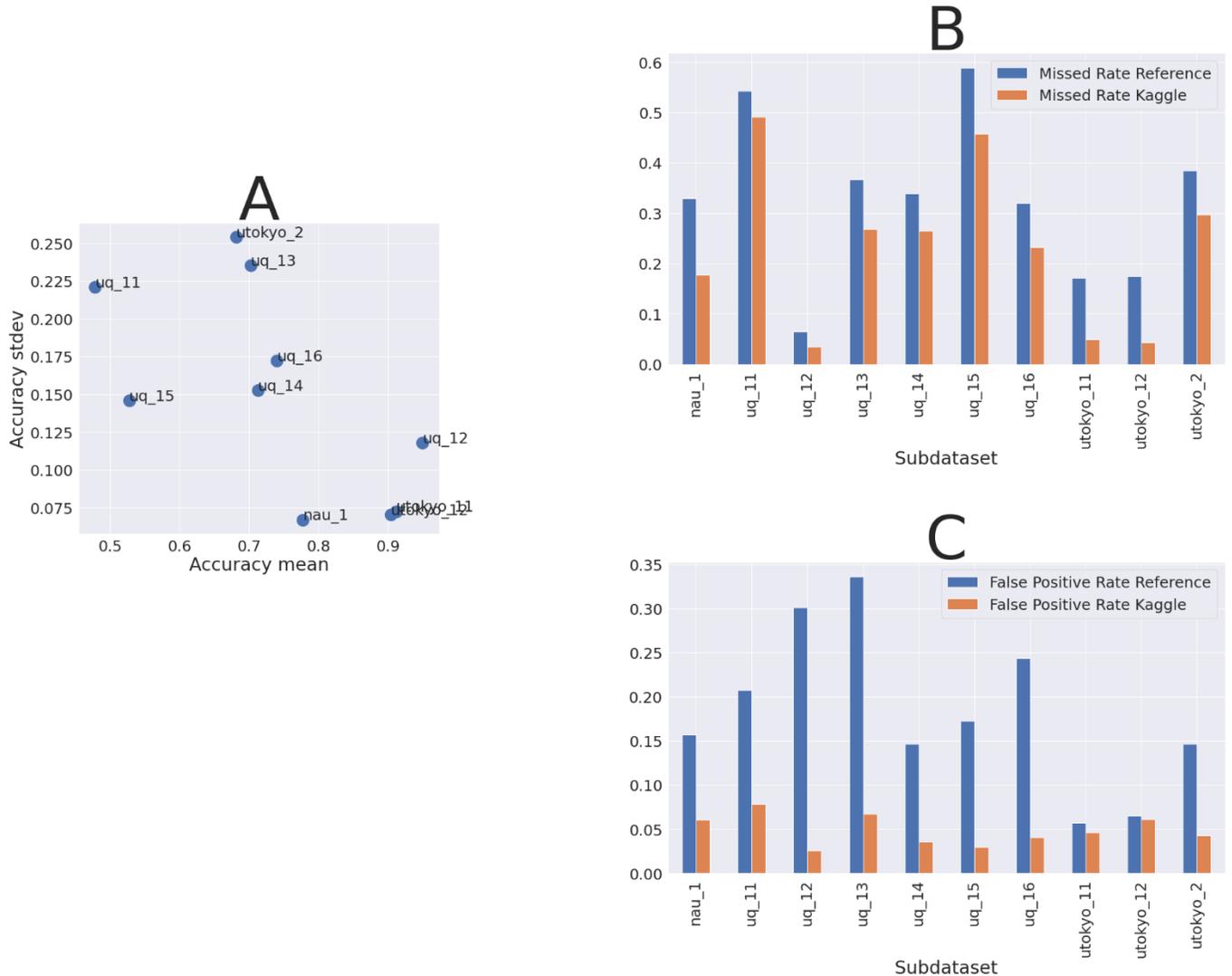

**Figure 4: Detailed performance of the best solution. Part a (on the left) details the mean accuracy for each domain in x-axis and the standard deviation in y-axis. An ANOVA has been realized to determine that the score distribution for each domain is different. Part b (on the right) compares the missed rate per domain and the false positive rate per domain. In blue are the results for the reference method described in (David *et al.*, 2020) and in orange is the Kaggle solution.**

The Kaggle solution was analyzed on the combination of the public and private test set, with the new split introduced by Global Wheat Head Dataset 2021. Figure 3a presents the mean accuracy for each domain in the x-axis and the standard deviation in the y-axis. We can observe that the performance of the Kaggle solution is very heterogeneous depending on the domain. An ANOVA confirms that the mean of each domain is different. It can be explained by the inner difficulty of the domain used in the test set. In figure

4b the missed rate and false positive rate are presented, compared to a faster-RCNN baseline from (**David et al., 2020**). The Kaggle solution demonstrated to be much more performant against false positives, which can be as impressive as a 7x decrease for the uq_13 domain. It also decreases the missed wheat head but with a much lower intensity. The highest errors from figure 3A are correlated with the missed wheat head ratio, explaining part of the error.

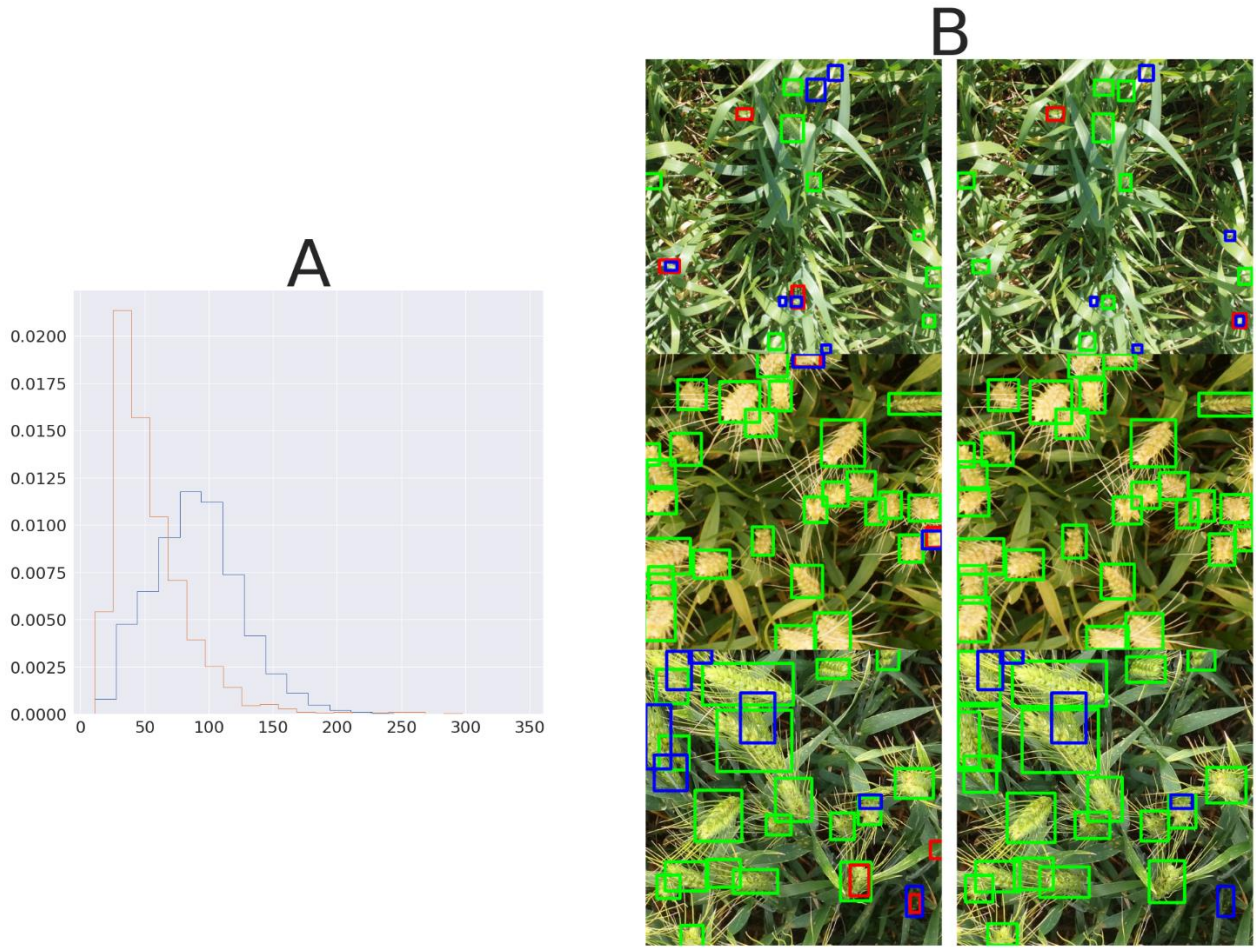

**Figure 5: Figure 5A : distribution of box size for the global test dataset in blue and the missed object size in orange. Figure 5B:Presentation of types of errors for the reference method and the Kaggle solution. The bounding boxes in green are the true positives, in blue are false negatives and in red are false positives.**

Figure 5 presents the results of evaluating random examples with the reference method and the Kaggle solution. The Kaggle solution demonstrates better results but, as discussed earlier, still struggles to detect objects in densely packed images. The high object density in these images results in overlapped wheat heads that are difficult to localize with detection methods that rely on a non-max suppression (NMS) algorithm. This is also linked with the detection of small wheat heads, which is a more difficult task. The size distribution of the missed object demonstrates that their size is often 2 times smaller than the usual wheat heads present in the dataset. The Kaggle solution appears to be a much more reliable than the baseline – detected wheat heads are much more likely to be true positive than a false positive. This is an interesting property for wheat head extraction and characterization. However, the missed rate is still high, resulting in an edge case where localization methods cannot be used for wheat head counting. Specific strategies to handle this edge case could be developed.

### 4.3. Analysis of Data Augmentation and pseudo-labelling performance

| Modality | Weighted Accuracy | Gain against baseline | Accuracy | Gain against baseline |
|---|---|---|---|---|
| Baseline | 0.452787 |  | 0.654569 |  |
| Baseline + Cutmix | 0.476214 | 0.023 | 0.659364 | 0.005 |
| Baseline + Mixup | 0.367481 | -0.085 | 0.553144 | -0.101 |
| Data Augmentation (DA) | 0.498065 | 0.045 | 0.704271 | 0.050 |
| DA + Cutmix | 0.501241 | 0.048 | 0.685546 | 0.031 |
| DA + Mixup | 0.437714 | -0.015 | 0.686517 | 0.032 |
| DA + Cutmix + Mixup | 0.358226 | -0.095 | 0.508334 | -0.146 |

Table 3: Ablation study different Data Augmentation strategies

The impact of Data augmentation is important to improve the quality of the robustness. The classic data augmentation increases the weighted accuracy by 0.045, almost 10% of the baseline score. Advanced data augmentation techniques such as Cutmix or Mixup do not seem beneficial for the training – while Cutmix alone improves performance by 0.023, it does not add any performance compared to classic data augmentation techniques. An interesting finding is that some data augmentation can decrease the robustness – the use of Mixup always decreases the performance compared to the baseline data. Our results suggest that while the use of Cutmix or Mixup theoretically increases robustness in the use case of classification, it does not seem to translate for detection; it yields marginal gains. Conclusions are similar when using the same accuracy as in the Kaggle challenge. The result is contradictory to the popularity of such approaches during the Kaggle competition.

In our experiment, Mixup could be drawn with a probability of 0.5, while it's reduced to 0.165 in the case of the winning solution. The quality of our classic data augmentation pipeline, which is inspired by the strongest solutions could also explain the results. Our results confirm the importance of Data Augmentation for robustness but call for more careful exploration when applying usual, typically multi-image, Data Augmentation techniques. It is particularly important to remember that results on classification tasks may not translate well on detection. The use of bounding boxes may also limit the use of strategies such as Cut, Paste and Mix to increase the diversity of the data. A potential axis of research would be to generate synthetic wheat heads to increase the diversity of the data. The use of GAN or Style Transfer may be a promising solution.

### 4.4. Extension to new domains

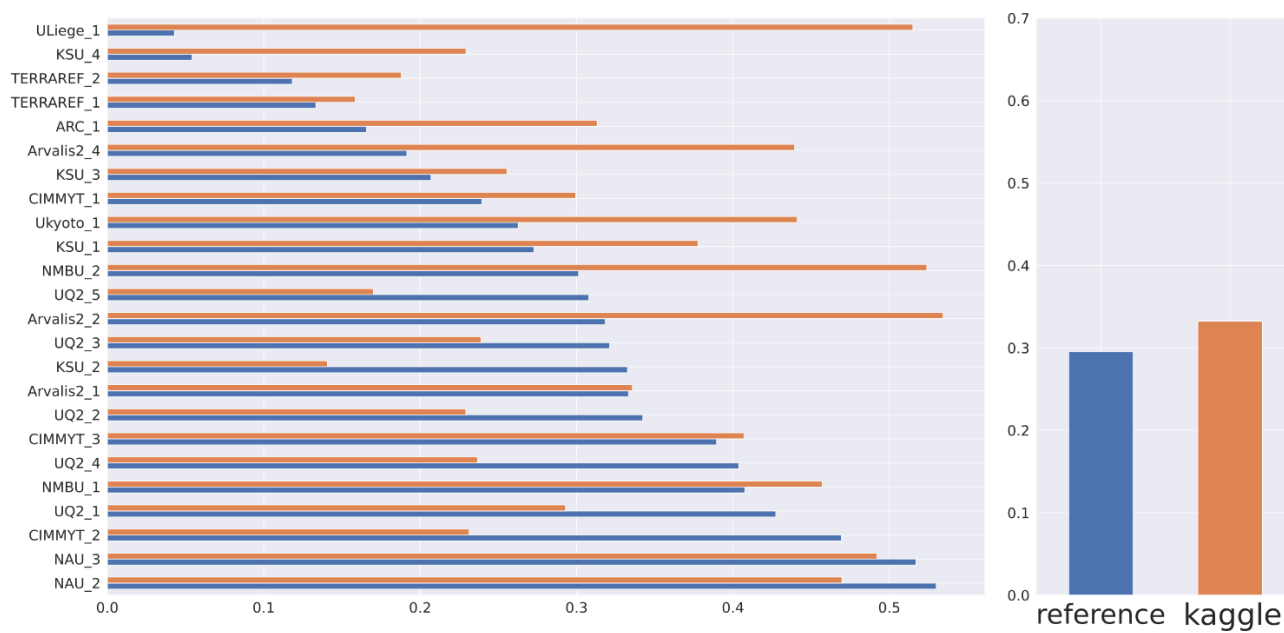

**Figure 6: Presentation of the results on the newly-added domains. Blue = Faster-RCNN, Orange = Weight from Kaggle solution.**

Results on the new test set are lower than on the older test set. This could be linked to the smaller size of the objects in the test set (but still in line with size of object in the training set). Some sub datasets are much more difficult due to more bias such as Terraref_1 and Terraref_2 (strong illuminations) or a lot of overlap due to the variety, stages and sowing density. The ULiege_1 and Arvalis2_4 datasets are in this category. On average, the Kaggle solution is more robust than the reference method but the margin is very tight and, for CIMMYT_2 and KSU_2, it is even lower than the reference.

The performance of the Kaggle solution does not extend to new solutions, even if it beats the reference method by 0.05. It overperforms in only 16 of the 27 sub datasets. Efforts in the competition has led to substantial improvements in the performance of wheat head localization, especially in the case of wheat heads with strong overlap. However, this performance boost came at the price of worse performance on some sub datasets. A part of the explanation is that small objects are a problem for current object detection models. Faster-RCNN and EfficientDet both rely on NMS to detect objects, which may lead to strong limitations for such competitions. An explanation for this disappointing finding may lie in the design of the competition. The metric used could have led to overfitting to the few domains contained in the test set. Interpretability of the solution improves with the additional test datasets. The number of four different domains for the 2020 competition and the non-averaged accuracy was a limitation that discouraged robust solutions. We hope that our findings can help the building of new competitions and benchmarks in Plant Phenotyping. We recommend using at least 10-15 different domains with contrasted conditions to get interpretable models.

## 5. Conclusion

The Global Wheat Challenge 2020 was an important step toward a general solution to wheat head detection from high resolution RGB imagery. First, it has attracted a lot of attention from competitors to a central problem in Plant Phenotyping and contributed to exposing the field to non-specialists, alongside other competitions such as Plant Seedling classification (Giselsson *et al.*, 2017), Plant Pathology classification (Thapa *et al.*, 2020) , Agriculture-vision challenge (Chiu *et al.*, 2020), or PlantClef (Goëau, Bonnet and Joly, 2019, 2020). It is unique in terms of the diversity of in-field situations and its focus on Plant Phenotyping. The competition can be used as a framework to rigorously test new approaches on object detection against previous works and help laboratories across the world to develop their own wheat head localization models. We hope that our findings can help other competition organizers to improve their design. Another competition will be held in 2021 with an extension of the dataset and a new metric derived from this study, which will be averaged between the domains. The diversity of the test dataset will be doubled compared to the 2020 competition.

A large range of approaches have been tested and documented, making the task of wheat head localization easier. However, the results of the study have revealed contradictions between the initial design and the expectations of the challenge, and the Kaggle competition may still not translate over to new contexts. Solutions that go beyond model optimization could have been missed. We encourage competitors to explore solutions linked to new architectures that can help to solve the problem of detecting small objects, especially when included in a large object, and to explore the literature for robust algorithms such as GDRO (Sagawa *et al.*, 2019).

## 6. Acknowledgements

The work received support from ANRT for the CIFRE grant of Etienne David, co-funded by Arvalis. The study was partly supported by several projects, including:

- Canada: the Global Institute Food Security, University of Saskatchewan supported the organization of the competition
- France :  PIA #Digitag Institut Convergences Agriculture Numérique , Hiphen supported the organization of the competition
- Japan: Kubota supported the organization of the competition
- Australia: Grains Research and Development Corporation (UOQ2002-008RTX Machine learning applied to high-throughput feature extraction from imagery to map spatial variability and UOQ2003-011RTX INVITA - A technology and analytics platform for improving variety selection) supported competition